\pdfoutput=1
\documentclass[1p, final]{elsarticle}

\usepackage{amssymb}
\usepackage{latexsym}
\usepackage{caption}
\usepackage{subcaption}
\usepackage{verbatim}
\usepackage{algorithm}
\usepackage{algorithmic}
\usepackage{tabularx}
\usepackage{amssymb}
\usepackage{amsmath}
\usepackage{amsthm}
\newcommand*{\ud} {{\,\mathrm{d}}}
\newcommand*\K {{\mathcal{K}}}

\newtheorem{proposition}{Proposition}
\newtheorem{corollary}{Corollary}

\journal{}

\begin{document}

\begin{frontmatter}

\title{Learning 2D Gabor Filters by Infinite Kernel Learning Regression}

\author[1]{Kamaledin Ghiasi-Shirazi\corref{cor1}} 
\cortext[cor1]{
  Tel.: +98-513-880-5158;  
  fax: +98-513-880-7181;
\ead{k.ghiasi@um.ac.ir}}
\address[1]{Department of Computer Engineering, Ferdowsi University of Mashhad (FUM), Office No.: BC-123, Azadi Sq., Mashhad, Khorasan Razavi, Iran.}

\begin{abstract}
Gabor functions have wide-spread applications in image processing and computer vision.  In this paper, we prove that 2D Gabor functions are translation-invariant positive-definite kernels and propose a novel formulation for the problem of image representation with Gabor functions based on infinite kernel learning regression. Using this formulation, we obtain a support vector expansion of an image based on a mixture of Gabor functions. The problem with this representation is that all Gabor functions are present at all support vector pixels. Applying LASSO to this support vector expansion, we obtain a sparse representation in which each Gabor function is positioned at a very small set of pixels. As an application, we introduce a method for learning a dataset-specific set of Gabor filters that can be used subsequently for feature extraction. Our experiments show that use of the learned Gabor filters improves the recognition accuracy of a recently introduced face recognition algorithm.
\end{abstract}

\begin{keyword}
Gabor kernels\sep stabilized infinite kernel learning \sep support vector regression
\end{keyword}

\end{frontmatter}

\section{Introduction}
Gabor functions are extensively used for feature extraction in numerous computer vision applications such as face recognition\citep{Cament20153371, ren2014band, Zhang20131, liu2003independent, yang2013gabor}, image retrieval \citep{Li2017118}, palmprint recognition\citep{Pan20092040, Pan20083032}, forgery detection\citep{Lee2015320}, and facial expression recognition\citep{Zhang2014451, Gu201280}.  
Neurologists have shown that receptive field of simple cortical cells can be modeled with Gabor functions\citep{daugman1980two, daugman1985uncertainty, marcelja1980mathematical}. 
For this reason, computational models for visual recognition that have inspired from visual cortex use Gabor filters in the very beginning phase of feature extraction \citep{serre2007robust}. 
Even nowadays that deep learning methods have remarkably influenced the field of machine vision, the combination of Gabor functions and convolutional neural networks(CNN) has been observed to improve recognition rates\citep{chang2014robust}. 

In this paper, we show that Gabor functions are translation-invariant (also called stationary) positive-definite kernels. 
It is somewhat strange that this fact, despite its simple proof, has been invisible to the eyes of researchers and it is neither mentioned in classical books such as \citep{Scholkopf_2002,Cristianini_2000, Taylor_2004} nor in seminal work of \citet{Genton_2001} who reviewed the class of stationary kernels\footnote{It must be mentioned that the term "Gabor kernel" is used in computer vision literature as a synonym to "Gabor filters" and does not refer to positive-definite kernels.}. 
We believe that the positive-definiteness of Gabor functions can potentially be exploited in numerous ways for applying kernel algorithms to machine vision problems. 
In this paper, we target the problem of learning Gabor filters from data which in kernel methods terminology is a kernel learning problem.
Perhaps the most widespread kernel learning algorithm is the multiple kernel learning (MKL) framework that seeks the best convex combination of a finite set of kernel functions\citep{Lanckriet_2004, Bach_2004, Sonnenburg_2006, Rakotomamonjy_2008, xu2009extended, kloft2011lp, bucak2014multiple}.  
One weakness of the MKL framework is that the initial set of kernel functions should be chosen by hand. 
To overcome this limitation, the infinite kernel learning (IKL) framework was introduced in which the set of initial kernels is extended to an infinite number of kernels parameterized over a continuous space \citep{Micchelli_2005, Argyriou_2005, Argyriou_2006, Gehler_2008, ozougur2010numerical, ozougur2010infinite, ghiasi2010learning}.
Some of the solutions proposed for this problem were restricted to Gaussian kernels \citep{Micchelli_2005, Argyriou_2005, Argyriou_2006} and some were restricted to binary classification with support vector machines \citep{Gehler_2008, ozougur2010numerical, ozougur2010infinite, ghiasi2010learning}. However, to apply these IKL algorithms to the problem of learning Gabor functions, one should formulate the problem of learning Gabor functions as binary classification which seems to be impossible. 
Fortunately, \citet{GhiasiShirazi2011} generalized the SIKL algorithm\citep{ghiasi2010learning} to a more general class of machine learning problems that includes the $\epsilon$-insensitive support vector regression (SVR). 
In this paper, we reduce the problem of representing an image with Gabor functions to the problem of learning a convex combination of an infinite number of Gabor kernels for regression.
This gives us a mixture of Gabor functions that, when placed at positions determined by support vectors, reconstruct the given image.
As a practical application of the SIKL algorithm, we propose a simple method for learning Gabor functions for a specific dataset of images from a tiny fraction of its images.
However, the representation obtained by the SIKL algorithm has the problem that all Gabor functions are present at all support vector pixels.
This may arouse the suspicion that the SIKL algorithm learns a universal approximator kernel function that is subsequently used by SVR for representing the input image, rejecting any link between the Gabor functions generating an image and the learned Gabor functions.
In fact, we will show experimentally that the mixture of Gabor functions learned by the SIKL algorithm is approximately a highly concentrated Laplacian kernel.
Using LASSO algorithm \citep{tibshirani1996regression}, we obtain a sparse representation of the original image in which the Gabor functions are located at a very sparse set of pixels. Experimental results on artificial images generated by combination of two Gabor functions confirm the potential of our sparse representation algorithm in discovering the scales, orientations, and locations of the constituting Gabor functions.
  
In Section~\ref{sec:sikl-review}, we give a concise and simplified introduction to SIKL regression \citep[the general form of SIKL algorithm and its mathematical analysis can be found in][]{GhiasiShirazi2011}.
  We introduce our method for representing an image as a mixture of Gabor functions in Section~\ref{sec:gabor-image-rep}.  
Our algorithm for choosing the parameters of Gabor filters for a specific dataset is given in Section~\ref{sec:learning-gabors-for-datasets}.
In Section~\ref{sec:sparse-representation}, we show how LASSO can be utilized to obtain a sparse Gabor-based representation of an image.
We experimentally evaluate the proposed method in Section~\ref{sec:experiments} and conclude the paper in Section~\ref{sec:conclusion}. In \ref{sec:gabor-form}, we give a formal proof for positive-definiteness of Gabor functions.

\section{Stabilized infinite kernel learning regression}\label{sec:sikl-review}

The stabilized infinite kernel learning (SIKL) algorithm had been initially introduced in \citep{ghiasi2010learning} for binary classification and then was generalized to more general classes of machine learning problems in \citep{GhiasiShirazi2011}.  
In this section, we give a short introduction to SIKL regression in a simple and succinct way without going into mathematical details and without grounding the SIKL framework in its most general form. For a comprehensive introduction to the SIKL framework the reader is referred to  \citep{GhiasiShirazi2011}.

Assume that the training set consists of the input samples 
$x_1,...,x_l\in\mathbb{R}^d$ and their corresponding target values
$y_1,...,y_l\in\mathbb{R}$.
Support vector regression attempts to learn the relation between input and output by a function of the form:

\begin{eqnarray}
f(x)=\sum_{i=1}^l(\hat{\beta}_i-\beta_i)k(x_i,x)+b.
\end{eqnarray}

The coefficients
$\hat{\beta}$ and $\beta$
are obtained by solving the following optimization problem
\citep{Cristianini_2000}:

\parbox{0cm}{
\begin{eqnarray*}
& \max_{\hat{\beta},\beta} & \sum_{i=1}^l y_i(\hat{\beta}_i-\beta_i) - \epsilon \sum_{i=1}^l(\hat{\beta}_i+\beta_i) \\
& &-\frac{1}{2}\sum_{i=1}^l\sum_{j=1}^l (\hat{\beta}_i-\beta_i) (\hat{\beta}_j-\beta_j) k(x_i,x_j)\\
& s.t. 	& 0 \leq \hat{\beta}_i, \beta_i \leq C  \;\;\; \textsf{for}\;\;\; i=1,...,l \\
& &  \sum_{i=1}^l (\hat{\beta}_i-\beta_i) = 0 
\end{eqnarray*}
}
\parbox{0.95\columnwidth}{
\begin{eqnarray} \label{eq:SvmRegressionFormulation}
\end{eqnarray}
}

\noindent
where $C\in(0,\infty]$ is a regularization constant.
The above optimization problem can be rewritten in the following more succinct form:

\begin{eqnarray}\label{eq:AssumedGeneralMachineLearningOptimization}
\max_{\alpha \in \mathcal{A}} -\frac{1}{2}\alpha^T A^T K A\alpha - c^T\alpha
\end{eqnarray}

\noindent
where $K$ is the kernel matrix obtained by applying the kernel function $k$ to the input samples $x_1,...,x_l$,

\begin{eqnarray}
\begin{aligned}
& \alpha=
\begin{bmatrix}
    \hat{\beta} \\
    \beta
\end{bmatrix}, \\
& A=
\begin{bmatrix}
    I_{\ell} \\ -I_{\ell}
\end{bmatrix}^T, \\
& c=
\begin{bmatrix}
	y_{1}\;\;
	y_{2}\;\;	
	\hdots  \;\;
   	y_{\ell}\;\;
	-y_{1}\;\;
	-y_{2}\;\;	
	\hdots  \;\;
   	-y_{\ell}
\end{bmatrix}^T -\varepsilon, \\
& \mathcal{A}=\left\{\alpha | 
\alpha=
\begin{bmatrix}
    \hat{\beta} \\
    \beta 
\end{bmatrix}, 
\; 0 \leq \alpha_i \leq C, \;
 \sum_{i=1}^\ell (\hat{\beta}_i-\beta_i) = 0\right\},
\end{aligned}
\end{eqnarray}

\noindent
and $I_{\ell}$ is the identity matrix of size $\ell$.

Consider the set of kernels $\left\{ k_\gamma: \gamma \in \Gamma \right\}$, where $\Gamma$ is a continuously parameterized index set. Let $\mathcal{P}(\Gamma)$ be the set of all probability measures on $\Gamma$. It can be shown \citep[see][]{Micchelli_2005} that for any probability measure $p\in \mathcal{P}(\Gamma)$, the function

\begin{eqnarray}\label{eq:convex-comb}
\hat{k}_p(x,z)=\int_{\Gamma} k_\gamma(x,z) \ud p(\gamma)
\end{eqnarray}

\noindent
is a convex combination of the set of kernels $\left\{ k_\gamma: \gamma \in \Gamma \right\}$. Conversely, any convex combination of the set of kernels $\left\{ k_\gamma: \gamma \in \Gamma \right\}$ can be written in the form of Eq.~(\ref{eq:convex-comb}). 
In the IKL framework, it is assumed that $\Gamma$ is a compact Hausdorff space (e.g. a bounded and closed subset of $\mathbb{R}^2$) and the problem is to find the best kernel in the form of Eq.~(\ref{eq:convex-comb}). 
The SIKL framework relaxes the assumption on $\Gamma$ to locally-compact Hausdorff spaces (e.g. $\mathbb{R}$ or $\mathbb{R}^2$). For mathematical concreteness and for provisioning a mechanism to control the capacity of the learning machine, the SIKL framework introduces a vanishing function\footnote{For metric spaces like $\mathbb{R}$ and $\mathbb{R}^2$ this means that the function tends to zero at infinity. 
For introduction to vanishing functions in general topological spaces please see \cite[page 70 of][]{Rudin:Real&ComplexAnalysis}.}
$G(\gamma):\Gamma\rightarrow[0,1]$ into the framework. The stabilized convex combination of the kernels
$\left\{ k_\gamma: \gamma \in \Gamma \right\}$
with stabilizer $G(\gamma)$ and probability measure 
$p\in\mathcal{P}(\Gamma)$ is defined as:

\begin{eqnarray}
\bar{k}_p(x,z)=\int_{\Gamma} k_\gamma(x,z) G(\gamma) \ud p(\gamma).
\end{eqnarray}

Correspondingly, the set of stabilized convex combination of kernels
$\left\{ k_\gamma: \gamma \in \Gamma \right\}$
with stabilizer $G(\gamma)$ is defined as:

\begin{eqnarray}
\K:=\left\{ \bar{k}_p : p\in \mathcal{P}(\Gamma) \right\}.
\end{eqnarray}

The problem of simultaneously learning the regression function along with a kernel function $\bar{k}_p\in\K$ can be formulated as:

\begin{eqnarray} \label{eq:SiklMainProblem}
\sup_{p \in \mathcal{P}(\Gamma)}\min_{\alpha \in \mathcal{A}} \left\{ \int_{\gamma\in\Gamma} \left[G(\gamma)\alpha^T A^T K(\gamma) A\alpha + c^T\alpha\right] \ud p(\gamma) \right\}
\end{eqnarray}

\noindent
where $K(\gamma)$ is the kernel matrix that is obtained by applying the kernel function $k_\gamma$ to the input training data $x_1,...,x_l$.

\citet{GhiasiShirazi2011} proved that the probability measure that optimizes the above problem is discrete with finite support. The SIKL toolbox optimizes the above problem by semi-infinite programming and returns the weights $\mu_i$ and the parameters $\gamma_i\in\Gamma$ which identify the optimal kernel $\bar{k}$ by the following formula:

\begin{eqnarray}
\bar{k}(x,z) = \sum_{i=1}^m \mu_i k_{\gamma_i}(x,z).
\end{eqnarray}

We added the Gabor kernel to the SIKL toolbox and exploited some special properties of Gabor kernels to optimize the toolbox. Specifically, since Gabor kernels are two-dimensional, we modified the global optimization algorithm of SIKL to search the space of parameters systematically.

\section{Gabor-based image representation using SIKL}\label{sec:gabor-image-rep}
In this section, we show how SIKL  regression can be applied to the task of image representation by Gabor functions.
We consider the following form for Gabor functions which is essentially a slightly modified version of the from chosen by \citep{ren2014band}:

\begin{eqnarray}\label{eq:gabor-our-form}
\begin{aligned}
\Psi &(x,y;x_0,y_0,\omega,\theta) = \\
& e^{-\frac{\omega^2 ((x-x_0)^2+(y-y_0)^2 )}{8\pi^2 }}  
 cos{\left(\omega ((x-x_0) cos{\theta}+(y-y_0) sin{\theta} )\right)}
\end{aligned}
\end{eqnarray}

\noindent
where the point $(x_0,y_0)$ is the center of the Gabor function in the spatial domain and the parameters $\omega$ and $\theta$ determine the scale and orientation of Gabor function, respectively. Note that, in Eq.~(\ref{eq:gabor-our-form}), the only inputs are $x$ and $y$ and $x_0$ and $y_0$ are parameters of the Gabor function. 
By considering $x_0$ and $y_0$ as inputs, we arrive at the following definition for Gabor kernels:

\begin{eqnarray}
\begin{aligned}
k_{\omega,\theta}&([x,y],[x',y']) =\Psi(x,y;x',y',\omega,\theta). 
\end{aligned}
\end{eqnarray}

There is another parameterization for Gabor functions which is obtained from Eq.~(\ref{eq:gabor-our-form}) by setting $\omega=\frac{\pi/2}{2^{\nu/2}}$ and $\theta = \mu \frac{\pi}{8}$. This $\mu\nu$-parameterization is specially important since manual selection of Gabor parameters is usually done in that form. We use this form when a parameter is to be chosen by hand or when reporting the learned parameters of Gabor functions.
\ref{sec:gabor-form} elaborates on the chosen form for Gabor functions and gives a proof for positive-definiteness of Gabor kernels.

Now, assume that we want to search for the best convex combination of Gabor kernels whose scale parameters are in the range 
$[\omega_{\ell 1},\omega_{u 1}]$. This choice corresponds to a rectangular vanishing function in SIKL formulation which is not appropriate due to the jumps from $0$ to $1$ and vise versa. Therefore, we choose the following trapezoidal  stabilizing function:

\begin{eqnarray}
	G(\omega,\theta)=\left\{
	\begin{array}{llll}
    &0 & \quad \quad & \omega < \omega_{\ell 0}\\
    &\frac{\omega-\omega_{\ell 0}}{\omega_{\ell 1}-\omega_{\ell 0}} & \quad \quad & \omega_{\ell 0}<\omega < \omega_{\ell 1}\\
    &1 & \quad \quad & \omega_{\ell 1}<\omega < \omega_{u 1}\\    
    &\frac{\omega-\omega_{u 0}}{\omega_{u 1}-\omega_{u 0}} & \quad \quad & \omega_{u 1}<\omega < \omega_{u 0}\\    
    &0 & \quad \quad & \omega > \omega_{u 0}\\    
    \end{array}
	\right.
\end{eqnarray}

The stabilized convex combination of Gabor functions with stabilizer $G(\omega, \theta)$ and probability measure $p$ is defined as:

\begin{eqnarray}
\begin{aligned}
\bar{k}_p &([x,y],[x',y'])\\
&=\int_{\mathbb{R}^2} k_{\omega ,\theta }([x,y],[x',y']) G(\omega ,\theta) \ud p(\omega ,\theta).
\end{aligned}
\end{eqnarray}

Consequently, the set of stabilized convex combination of Gabor functions with stabilizer $G(\omega, \theta)$ can be expressed as:

\begin{eqnarray}
\K:=\left\{ \bar{k}_p : p\in \mathcal{P}(\mathbb{R}^2) \right\}.
\end{eqnarray}

As stated previously, although the optimization is over a continuous space of parameters, the optimal kernel has a finite expansion of the form:

\begin{eqnarray}\label{eq:kernel-finite-form}
\bar{k}([x,y],[x',y'])=\sum_{i=1}^m \mu_i k_{\omega_i ,\theta_i } (x-x',y-y' ).
\end{eqnarray}

For a given image $I$, we generate a training set that consists of positions of pixels as input and the intensity at those pixels as desired outputs. We then use the SIKL regression algorithm to learn the above kernel and the parameters of a SVR machine simultaneously in order to predict the intensity of each pixel correctly.
The solution of the SIKL problem gives the number of participating kernels $m$, Gabor parameters $\omega_i$ and $\theta_i$ for $i=1,...,m$, and the support vector coefficients $\hat{\beta}_i,\beta_i$ for $i=1,...,\ell$, where $\ell$ is the number of pixels in the image, such that:

\begin{eqnarray}\label{eq:double-sigma-approx}
I(x,y) \approx  \sum_{i\in SV}{(\hat{\beta}_i-\beta _i ) \sum_{j=1}^m{\mu_j k_{\omega _j,\theta _j} (x-x_i,y-y_i)}} + b.
\end{eqnarray}

This representation signifies the Gabor functions that are contributing to the construction of the input image $I$.

\section{Learning dataset-specific Gabor filters}\label{sec:learning-gabors-for-datasets}
When Gabor filters are used for feature extraction from a dataset, their parameters are usually tuned by hand and it is customary to use 40 Gabor functions with 5 scales and 8 directions\citep{liu2002gabor,liu2004gabor,ren2014band, haghighat2013identification}. 
However, since Gabor functions are defined over a pixel-space, appropriate choice of their parameters is sensitive to the resolution of the images. 
In Section~\ref{sec:gabor-image-rep}, we proposed an algorithm for learning an image representation based on Gabor functions by SIKL. 
It is an accepted practice in machine learning that the first phases of information processing usually model the distribution of the input data  while the task of discrimination is assigned to higher layers \citep {bishop1995neural, erhan2010does}.
So, we take the assumption that the Gabor functions that are appropriate for representing  an image, can also be used for feature extraction. 
By clustering the parameters obtained from a small fraction of images from a dataset using the k-means algorithm, we obtain a set of Gabor functions that are appropriate for representing any image in that dataset. Dataset-specific details on our method for learning Gabor filters for CMU-PIE and EYaleB datasets are given in Section~\ref{sec:exp-face}.

\section{Sparse image representation using Gabor kernels}\label{sec:sparse-representation}
The Gabor kernels learned by the method proposed in the previous section are global in the sense that each kernel is present at every location. 
In Section~\ref{sec:exp-analysis} we show that the mixture of the learned Gabor functions is approximately a concentrated Laplacian kernel. 
It may be questioned whether the Gabor kernels learned by the SIKL algorithm are those that are actually participating in the generation of an image or the learned combined concentrated Laplacian kernel acts as a universal approximator function that can be utilized by the SVR machine for approximating any input image. 
In this section, we aim to represent an image sparsely by a combination of Gabor functions such that each Gabor function is located at a small number of pixels. 
It has the benefit that it associates Gabor functions to the specific locations at which they are present. 
This problem has been previously considered by \citet{fischer2006sparse} who proposed an algorithm based on local competition. 
It must be mentioned that the set of Gabor functions chosen by the SIKL algorithm is already sparse. This sparseness is the result of the implicit $L1$ constraint $\lVert p\rVert_1=1$ over the probability measure $p$ in Eq.~(\ref{eq:SiklMainProblem}) which holds since the Lebesgue integral of any probability measure is 1. 
Thus, we assume that all the Gabor kernels that are found by the SIKL algorithm should be present in the sparse representation as well. We then try to sparsify the set of pixels at which each kernel is present.
We start from Eq.~(\ref{eq:double-sigma-approx}) obtained in the previous section. By exchanging the order of summation we obtain:

\begin{eqnarray} \label{eq:double-sigma-approximation}
I(x,y)
\approx
\sum_{j=1}^m \sum_{i\in SV}{[\mu_j(\hat{\beta}_i-\beta _i )] k_{\omega _j,\theta _j} (x-x_i,y-y_i)} + b.
\end{eqnarray}

Our goal is to approximate the inner summation with a sparse combination of the training input data. Let $b^j$ be an $\ell\times 1$ vector whose n'th element is:

\begin{eqnarray}
b^j_n = \sum_{i\in SV}{[\mu_j(\hat{\beta}_i-\beta _i )] k_{\omega _j,\theta _j} (x_n-x_i,y_n-y_i)}.
\end{eqnarray}

Assume $K^j$ is the $\ell\times \#sv$ kernel matrix associated with the kernel function  $k_{\omega_j,\theta_j}$ in which rows correspond to the image coordinates and columns correspond to the support vector image coordinates. 
According to Eq.~(\ref{eq:double-sigma-approximation}), to obtain a sparse representation for image $I$, we should find a sparse vector $\rho^j$ such that,
for $n=1,...,\ell$, we have:

\begin{eqnarray}\label{eq:expanded-form}
\begin{aligned}
b^j_n &= \sum_{i\in SV}{[\mu_j(\hat{\beta}_i-\beta _i )] k_{\omega _j,\theta _j} (x_n-x_i,y_n-y_i)} \\
&\approx\sum_{i\in SV}{\rho^j_i k_{\omega _j,\theta _j} (x_n-x_i,y_n-y_i)}.
\end{aligned}
\end{eqnarray}

Eq.~(\ref{eq:expanded-form}) can be written in the matrix notation as:

\begin{eqnarray}\label{eq:bKrho}
b^j \approx K^j \rho^j.
\end{eqnarray}

We have:

\begin{eqnarray} 
\begin{aligned}
I(x_n,y_n)
&\approx
\sum_{j=1}^m \sum_{i\in SV}{[\mu_j(\hat{\beta}_i-\beta _i )] k_{\omega _j,\theta _j} (x_n-x_i,y_n-y_i)} + b \\
&= \sum_{j=1}^m {b^j_n} + b 
\approx \sum_{j=1}^m {K^j_{n*} \rho^j} + b \\
&= \sum_{j=1}^m \sum_{i\in SV}{\rho^j_i k_{\omega_j,\theta_j}(x_n-x_i,y_n-y_i)} + b 
\end{aligned}
\end{eqnarray}

\noindent
where $K^j_{n*}$ is the n'th row of the kernel matrix $K^j$ and sparseness of this representation follows from the sparseness of the vector $\rho^j$. To find a sparse vector $\rho^j$ that satisfies Eq.~(\ref{eq:bKrho}), we use LASSO\citep{tibshirani1996regression} which solves the following optimization problem:

\begin{eqnarray}\label{eq:lasso}
\min_{\rho^j}\left\{
\frac{1}{\ell} 
\left\lVert b^j - K^j\rho^j \right\lVert_2^2 + 
\lambda \left\lVert \rho^j\right\rVert_1 
\right\}.
\end{eqnarray}

To undo the negative effect of the $L1$ regularization term in LASSO on the quality of approximation of Eq.~(\ref{eq:bKrho}), we again solve Eq.~(\ref{eq:bKrho}) using the least squares method with the constraint that the pattern of sparseness of $\rho^j$ found by LASSO would be preserved.

\section{Experiments}\label{sec:experiments}
The experiments of this section are designed with two goals in mind. First, we want to analyze the proposed algorithm in details and discover the nature of the learned Gabor functions. Second, we want to show the usefulness of the proposed method in automatic learning of Gabor functions for a given dataset. 
In Section~\ref{sec:exp-face}, we report our experiments on the application of the learned Gabor functions to the face recognition problem and show that it yields favorable recognition accuracy over a hand-tuned choice made by experts.
In Section~\ref{sec:exp-analysis}, we analyze the learned Gabor functions and show that the weighted combination of learned Gabor kernels is equivalent to a concentrated Laplacian kernel. Finally, in Section~\ref{sec:exp-sparse}, we analyze the proposed algorithm for Gabor-based sparse representation of images.

\subsection{Selection of Gabor filter for face recognition}\label{sec:exp-face}
In this section, we want to show that use of Gabor filters learned by the method proposed in Section~\ref{sec:learning-gabors-for-datasets} can increase the accuracy of machine vision applications compared with Gabor filters chosen by hand. 
For this purpose, we chose the MOST system that is recently proposed by \citet{ren2014band} for the task of face recognition and uses Gabor filters for feature extraction. 
The code of the MOST algorithm along with the CMU-Light and EYaleB face datasets were obtained by contacting \citet{ren2014band}.
CMU-Light is the name \cite{ren2014band} gave to the illumination part of CMU-PIE dataset \citep{baker2003cmu} which consists of 43 images captured at different illumination conditions from 68 persons, amounting to 2924 images. 
The Extended Yale B dataset \citep{georghiades2001few}, which is abbreviated as EYaleB, consists of 64 frontal images from 38 persons again taken at different illumination conditions, amounting to 2432 images. \citet{ren2014band} removed the 5 most dark images from the original 64 instances provided for each person in Extended Yale B dataset.  
In addition, all images had been histogram-equalized and were resized to width 46 and height 56. 

\begin{figure}[h] 
        \centering
		\includegraphics[width=1\columnwidth]{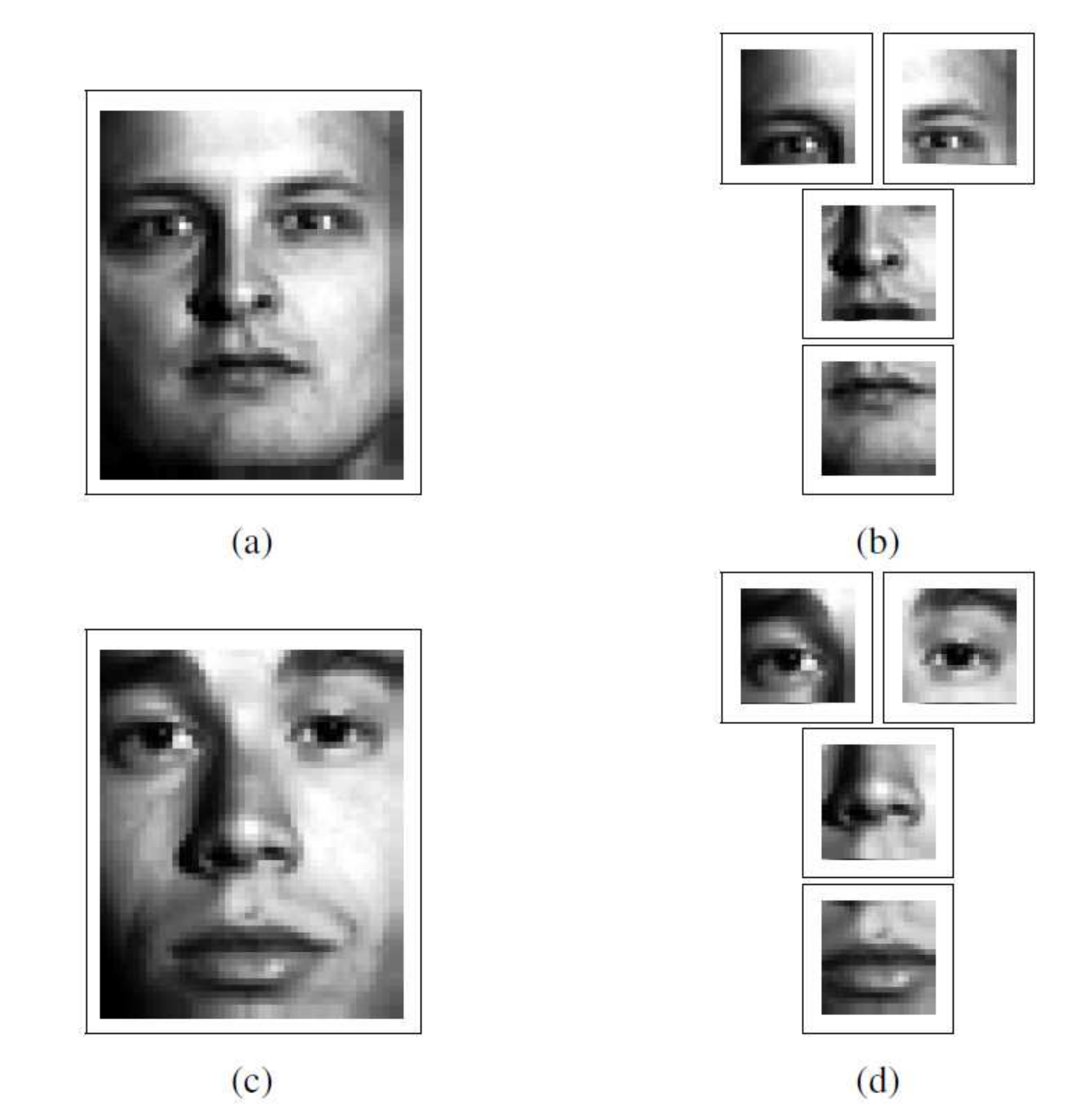}
  \caption{Regions for extracting subimages from face images. Note that two regions are overlapping. (a) and (b): An image from CMU-Light dataset and its associated subimages. (c) and (d): An image from EYaleB dataset and its associated subimages.}
  \label{fig:face-regions}
\end{figure}

Since the SIKL algorithm is time-consuming, considering the locality of Gabor functions, instead of representing a whole image with Gabor functions, we break images into several smaller (sometimes overlapping) regions and represent each region with a set of Gabor functions. 
From each face, we extract four regions around the two eyes, the nose, and the mouth (see Figure~\ref{fig:face-regions}). 
We used a trapezoidal vanishing function in the formulation of the SIKL algorithm with 
$\omega_{\ell 0}=\frac{\pi}{512}$,
$\omega_{\ell 1}=\frac{\pi\sqrt{2}}{512}$, 
$\omega_{u 1}=2\sqrt{2}\pi$, 
and $\omega_{u 0}=4\pi$. In $\mu\nu$-space, these choices correspond to  
$\nu_{\ell 0}=16$,
$\nu_{\ell 1}=15$, 
$\nu_{u 1}=-5$, 
and $\nu_{u 0}=-6$.

\begin{figure}[h] 
\centering
	\begin{subfigure}[t]{1\textwidth}
        \centering
\includegraphics[width=2.6in]{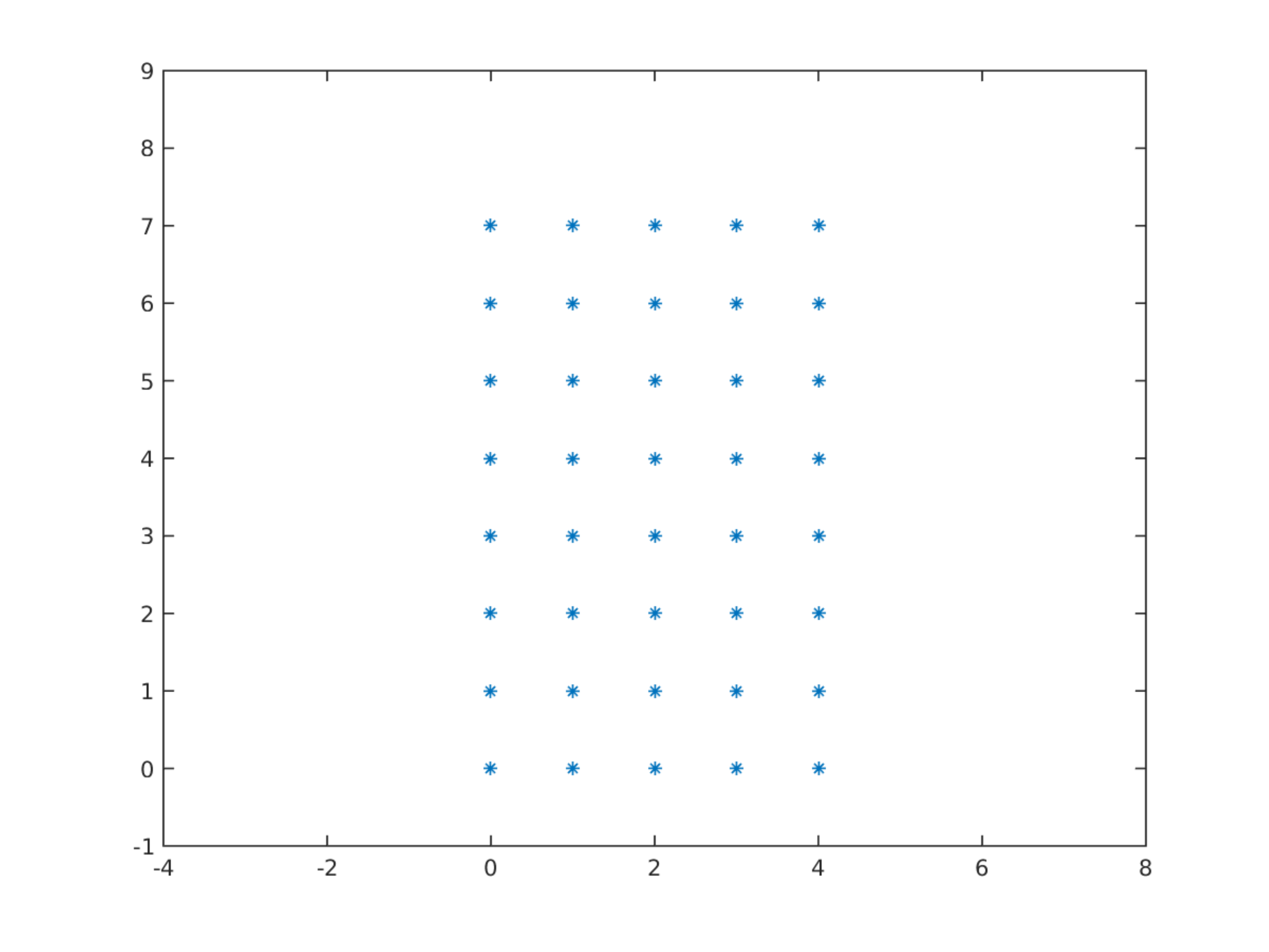}
\caption{Parameters of the original $40$ Gabor functions used by \citet{ren2014band}.}
    \end{subfigure}
	\vfill
	\begin{subfigure}[t]{1\textwidth}
        \centering
\includegraphics[width=2.6in]{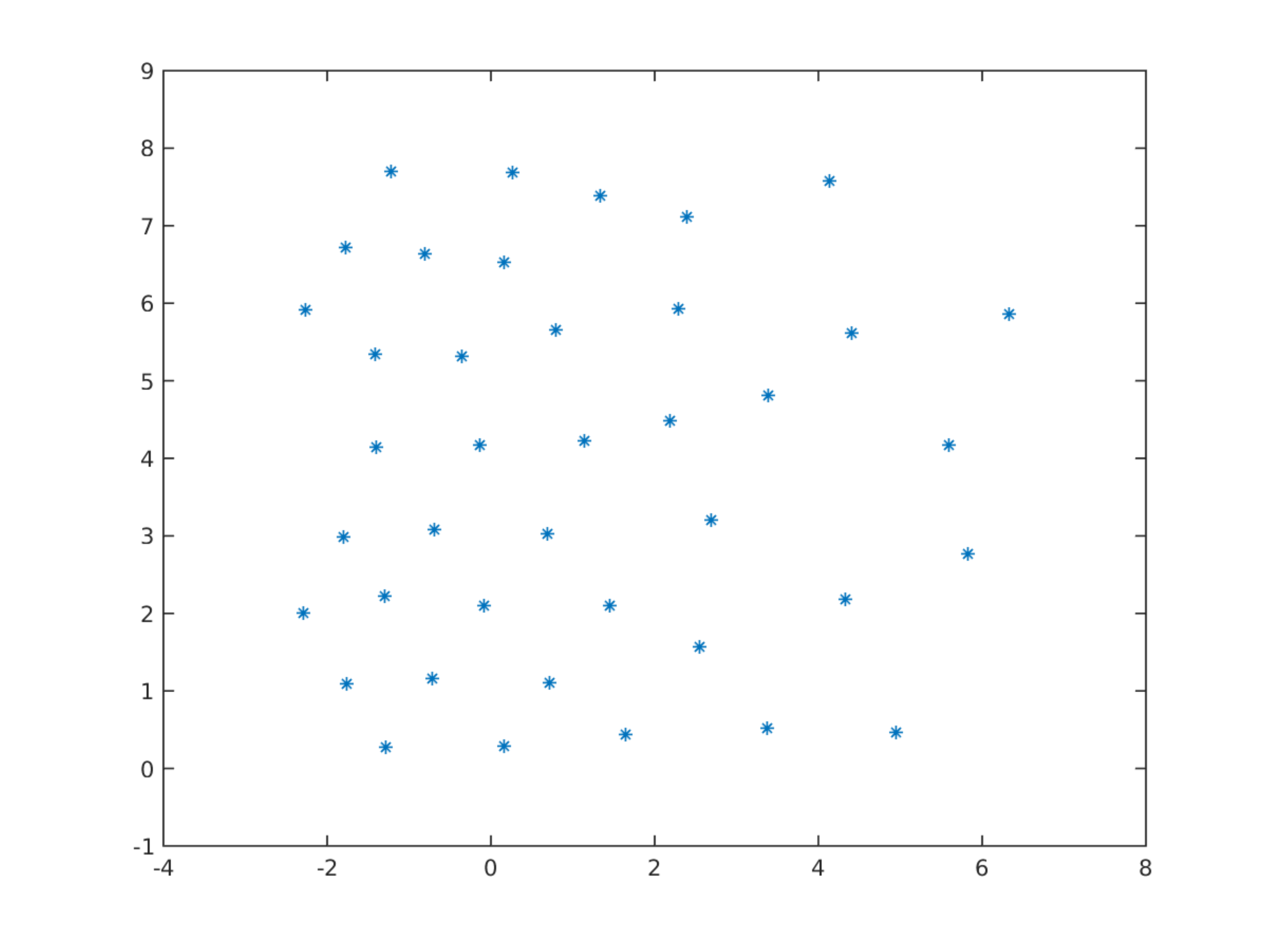}
\caption{Parameters of the $40$ learned Gabor functions for the CMU-Light dataset.}
    \end{subfigure}
	\vfill
	\begin{subfigure}[t]{1\textwidth}
        \centering
\includegraphics[width=2.6in]{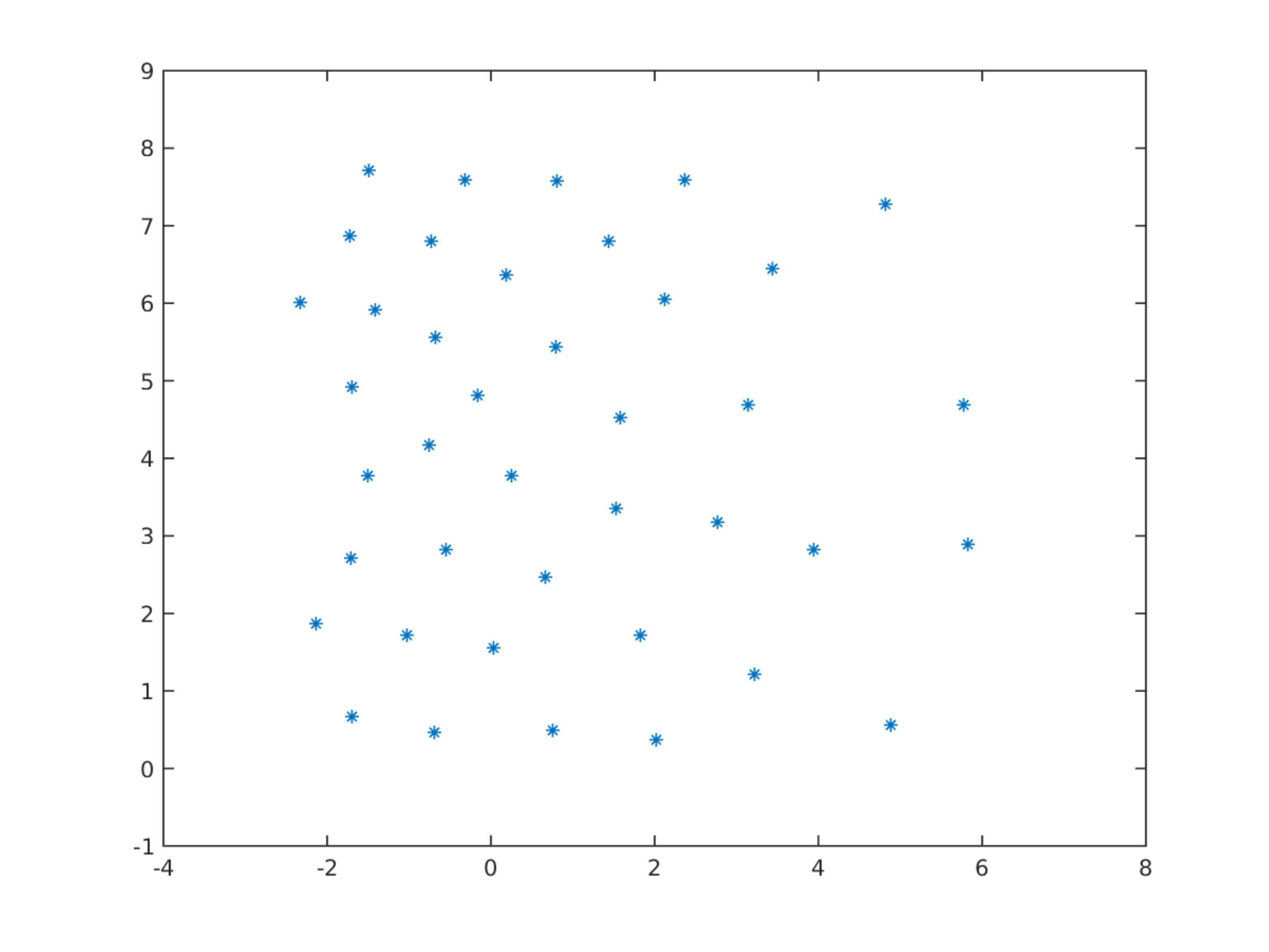}
\caption{Parameters of the $40$ learned Gabor functions for the EYaleB dataset.}
    \end{subfigure}
  \caption{Parameters of handy and learned Gabor functions depicted in $\mu\nu$-space. The horizontal axis shows the scale ($\nu$) and the vertical axis shows the orientation ($\mu$).}
  \label{fig:Gabor-parameters-uv-space}
\end{figure}

From each dataset, we randomly selected 28 images (which in both cases amounts to less than $2\%$ of the data) for learning the parameters of Gabor filters. 
We used the kmeans algorithm to cluster the Gabor parameters extracted from these $28$ images to obtain $40$ Gabor filters. 
Finally, we evaluated the original and learned Gabor filters on the task of face recognition using the MOST method. 
The number of training images used by the MOST algorithm, called \emph{ntrain}, is an important factor in the accuracy of the face recognition system. 
We compare the accuracies obtained by the original $40$ filters used by \citet{ren2014band} and the $40$ filters learned by our method. 
Each experiment is repeated $30$ times. 
The results of these experiments are summarized in Table~\ref{table:Gabor-filters}. 
As can be seen, when the number of training images for the MOST algorithm is low, use of the learned Gabor filters significantly increases the recognition rate. 
Figure~\ref{fig:Gabor-parameters-uv-space} shows the parameters of the original and the learned filters in the $\mu\nu$-space. 
It is clear from the figure that the parameters of Gabor filters used by \citet{ren2014band} do not cover the whole region of parameters that are indeed required for representing images with Gabor functions. 
In addition, the dataset-specific distributions for Gabor kernel parameters depicted in Figures~\ref{fig:Gabor-parameters-uv-space}.b and \ref{fig:Gabor-parameters-uv-space}.c can be used as a guideline for manual tuning of parameters of Gabor filter.

\begin{table}[h]
\caption{Comparison of accuracies obtained by MOST algorithm on CMU-Light and EYaleB datasets when using the manually tuned Gabor filters of \citet{ren2014band} and when using Gabor filters learned by the proposed algorithm. The parameter "ntrain refers to the number of training faces used by the MOST face recognition algorithm\citep{ren2014band}. The proposed method uses less than $2\%$ of images of each dataset for learning the parameters of Gabor kernels. For each dataset, the last column shows the two-tailed P-values for paired t-test. Results that are statistically significant are bold-faced. It must be emphasized that since a \emph{paired} t-test is used, P-values cannot be computed from statistics summarized in this table.}
\begin{center}
\begin{tabular}{|l|r|r|r|}\hline
& \multicolumn{3}{c}{CMU-Light} \vline \\
\hline
\multicolumn{1}{|l}{ntrain} \vline 
& \multicolumn{1}{l}{manually tuned} \vline
& \multicolumn{1}{l}{learned} \vline 
& \multicolumn{1}{l}{P-value} \vline\\
\hline
1	& $\mathbf{79.13\pm 5.7001}$	& $\mathbf{81.66 \pm  5.62}$	 & $\mathbf{<0.0001}$ \\
2	& $\mathbf{91.15\pm 5.8534}$	& $\mathbf{92.18 \pm  5.37}$	 & $\mathbf{0.0016}$ \\	
3	& $97.39\pm 4.0537$	& $97.54 \pm  3.74$	& $0.4522$ \\
4	& $98.95\pm 2.0855$	& $98.90 \pm  1.91$	& $0.2268$ \\
5	& $\mathbf{99.47\pm 1.1044}$	& $\mathbf{99.36 \pm  1.13}$	& $\mathbf{0.040}$ \\   
   \hline
\end{tabular}

\hfill \\
\hfill \\

\begin{tabular}{|l|r|r|r|}\hline
& \multicolumn{3}{c}{EYaleB}  \vline \\
\hline
\multicolumn{1}{|l}{ntrain} \vline 
& \multicolumn{1}{l}{manually tuned} \vline
& \multicolumn{1}{l}{learned} \vline 
& \multicolumn{1}{l}{P-value} \vline \\
\hline
1   & $\mathbf{37.65\pm 2.99}$	& $\mathbf{39.96 \pm  3.89}$	& $\mathbf{<0.0001}$ \\
2	& $\mathbf{72.99\pm 2.88}$	& $\mathbf{73.71 \pm  2.90}$	& $\mathbf{0.0089}$ \\	
3	& $88.22\pm 2.45$	& $88.46 \pm  2.03$	& $0.2840$	 \\
4	& $94.42\pm 1.72$	& $94.36 \pm  1.55$	 & $0.6407$\\
5	& $97.09\pm 1.00$	& $96.99 \pm  0.96$	& $0.3836$	 \\   
   \hline
\end{tabular}

\end{center}
\label{table:Gabor-filters}
\end{table}

\subsection{Analysis of the learned Gabor functions}\label{sec:exp-analysis}
\begin{figure}[h] 
\centering
	\begin{subfigure}[t]{0.45\textwidth}
        \centering
        \includegraphics[width=2.5in]{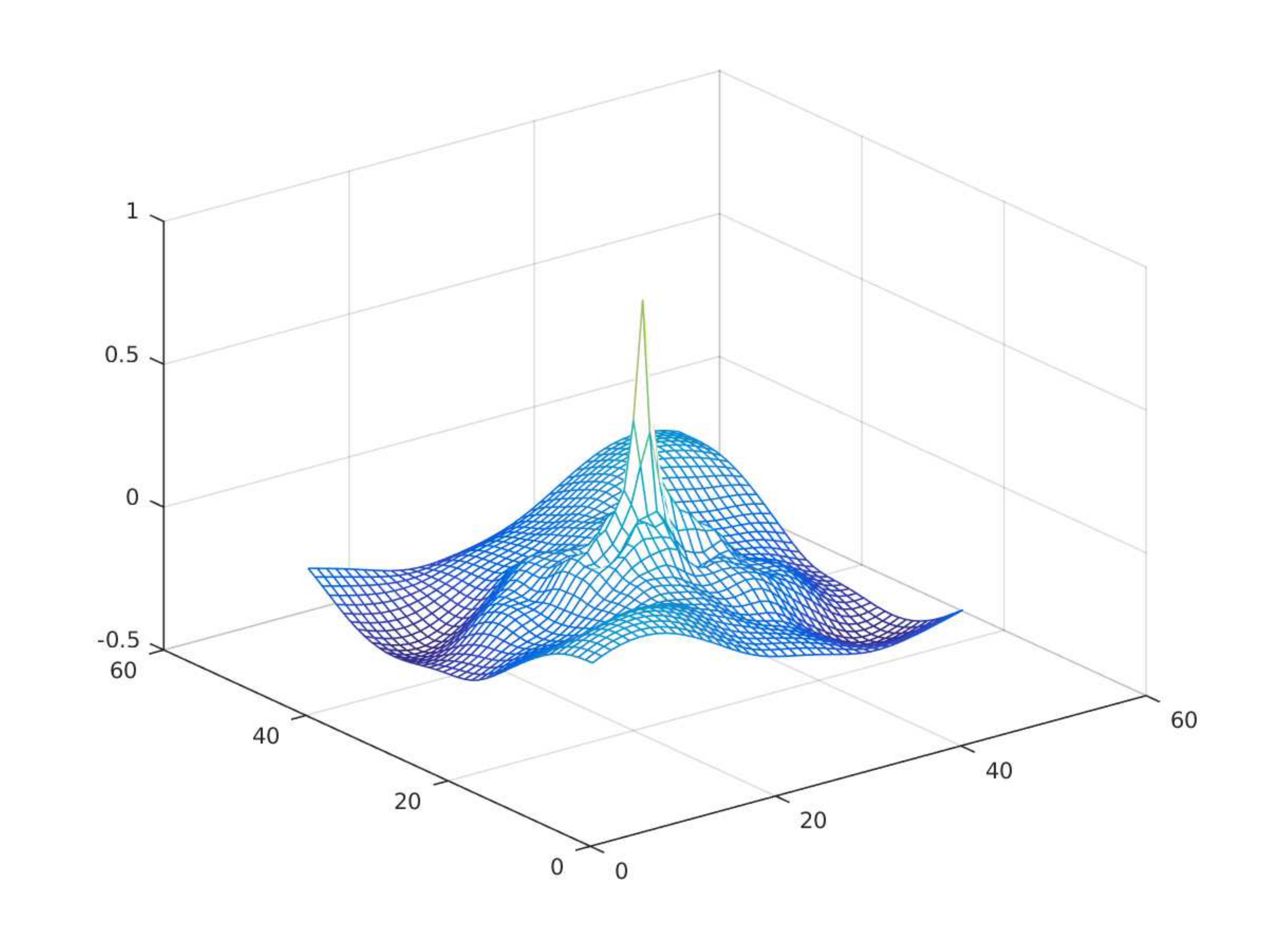}
		\caption{Learned combined kernel for a sample image region from CMU-Light dataset.}
    \end{subfigure}
	\hfill
	\begin{subfigure}[t]{0.45\textwidth}
        \centering
		\includegraphics[width=2.5in]{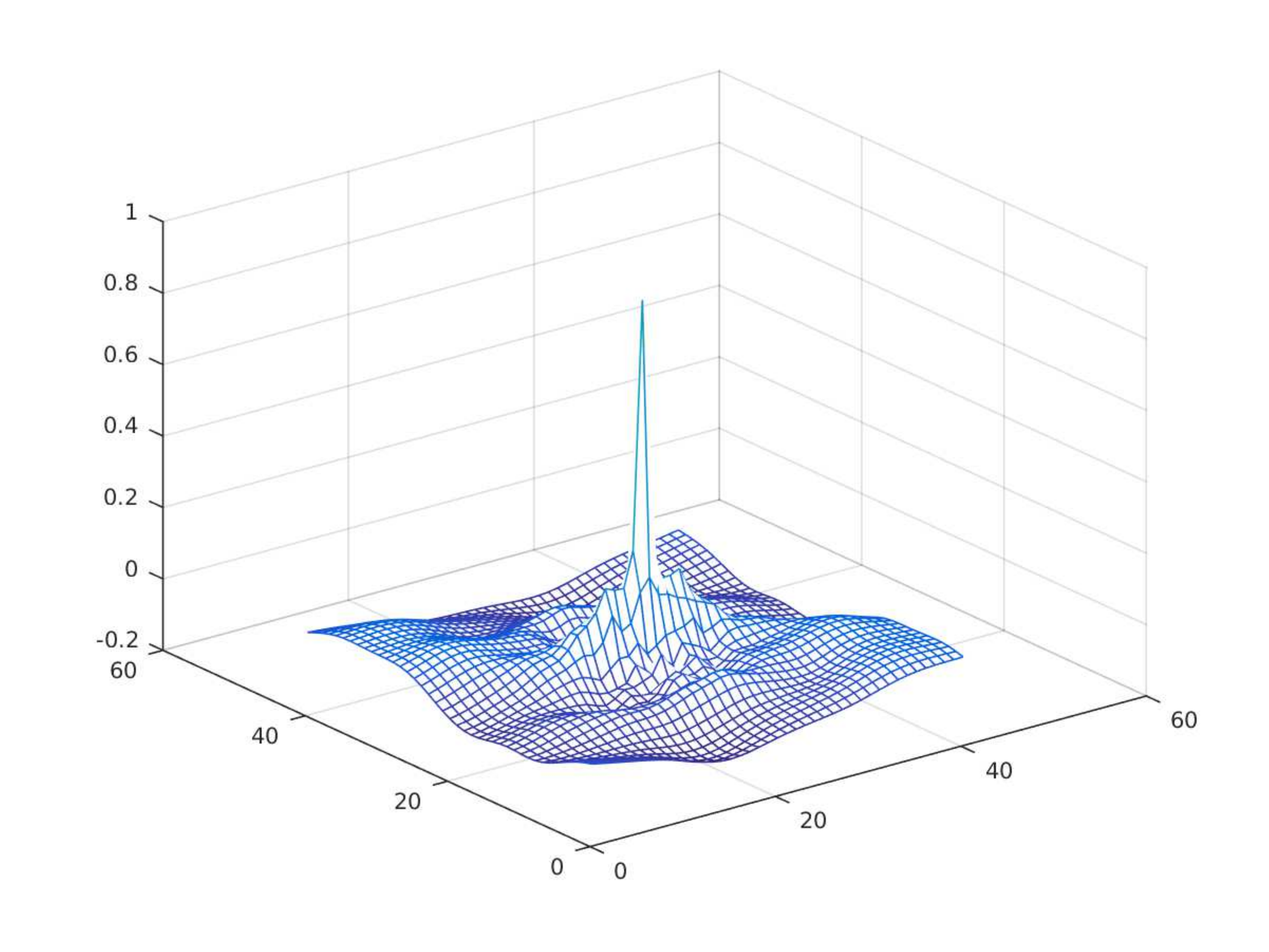}
		\caption{Learned combined kernel for a sample image region from EYaleB dataset.}	
    \end{subfigure}
 \caption{Compound kernels corresponding to the weighted combination of learned Gabor kernels by the method of Section~\ref{sec:gabor-image-rep}.}
  \label{fig:learned-kernels}
\end{figure}

In Section~\ref{sec:gabor-image-rep}, we showed how the SIKL algorithm can be exploited for representing an image with Gabor kernels.
The learned representation can be equivalently obtained by support vector regression with a single kernel that is the weighted combination of the selected Gabor kernels (see Eq.~\ref{eq:kernel-finite-form}).
An interesting question is what is the single kernel that is equivalent to the combination of the learned Gabor functions. 
We answer this question empirically by drawing the shape of the combined kernel. 
Figure~\ref{fig:learned-kernels} shows the combined kernels for two sample images from CMU-Light and EYaleB datasets. 
As can be seen, the weighted combination of the learned Gabor functions is approximately a concentrated Laplacian kernel.

\subsection{Discovering locations of constituting Gabor functions}\label{sec:exp-sparse}
\begin{figure}[h!] 
\centering
	\begin{subfigure}[t]{\columnwidth}
        \centering
        \includegraphics[width=3in]	{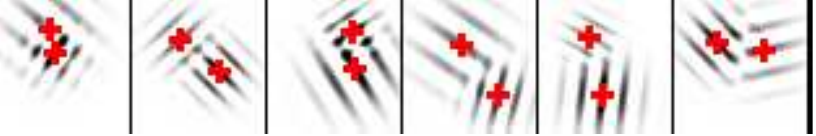}
		\caption{}
    \end{subfigure}
	\hfill
	\begin{subfigure}[t]{1\columnwidth}
        \centering
        \includegraphics[width=3in]	{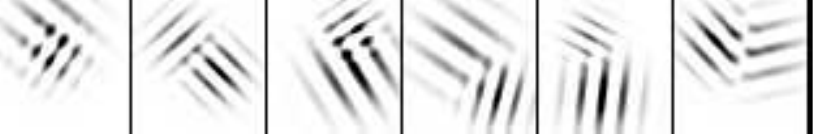}
		\caption{}
    \end{subfigure}
	\hfill
	\begin{subfigure}[t]{1\columnwidth}
        \centering
        \includegraphics[width=3in]	{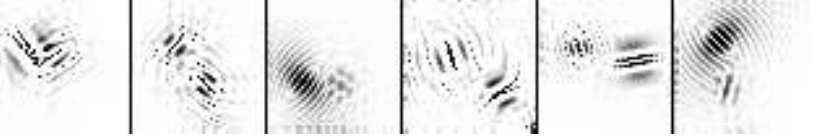}
		\caption{}
    \end{subfigure}
	\hfill		
	\begin{subfigure}[t]{1\columnwidth}
        \centering
		\includegraphics[width=3in]{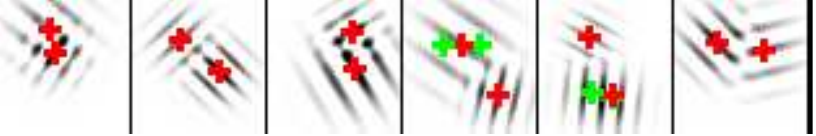}
		\caption{}	
    \end{subfigure}
 \caption{(a) Some randomly generated images by combination of two Gabor functions. Centers of Gabor functions are marked with a red plus sign. 
 (b) Non-sparse approximation of images of subfigure (a) with SIKL algorithm. Since the learned Gabor functions are present at all support vector locations, they cannot be assigned to any specific location.
 (c) An image of support vector pixels in which the darkness of each pixel is proportional to the magnitude of its support vector coefficient.  
 (d) Approximation of images of subfigure (a) obtained by learning a sparse Gabor-based representation. Red/Green plus signs correspond to the centers of the Gabor functions contributing positively/negatively in the learned sparse representation.}
  \label{fig:sparse-representaion}
\end{figure}

In this section, we experimentally evaluate the SIKL+LASSO algorithm of Section~\ref{sec:sparse-representation} in discovering the exact location of Gabor functions participating in a sparse representation of an image.
For this purpose, we first produced a few artificial images by combination of two randomly generated Gabor functions. Figure~\ref{fig:sparse-representaion}.a shows several examples of these artificially generated images.
Then, we used the SIKL+LASSO method proposed in Section~\ref{sec:sparse-representation} for discovering the positions of the original Gabor functions. 
We used a regularization constant of $\lambda=0.1$ for the LASSO algorithm.
Figure~\ref{fig:sparse-representaion}.b shows the approximations of images of Figure~\ref{fig:sparse-representaion}.a  generated by the SIKL algorithm.
The set of support vector pixels found by the SIKL algorithm are depicted in Figure~\ref{fig:sparse-representaion}.c. 
The approximations of the images of Figure~\ref{fig:sparse-representaion}.a  generated by the SIKL+LASSO algorithm along with the positions of the discovered Gabor functions are shown in Figure~\ref{fig:sparse-representaion}.d.
As can be seen, both the SIKL algorithm of Section~\ref{sec:gabor-image-rep} and the SIKL+LASSO algorithm of Section~\ref{sec:sparse-representation} generate acceptable approximations to the original images. 
On the other hand, while the set of support vectors obtained by  the SIKL algorithm contains many pixels, the SIKL+LASSO algorithm has been successful in obtaining a very sparse representation of the images.
However, in some cases the Gabor functions learned by the SIKL+LASSO algorithm do not correspond exactly to the generating ones.
It must be mentioned that since Gabor functions constitute an overcomplete system, it is natural that an image can be represented by different combinations of these functions. 
Noting that the SIKL+LASSO method uses exactly  those Gabor kernels that had been obtained by the SIKL method, this experiment reveals that the set of Gabor kernels learned by the SIKL algorithm is strongly related to those generating an image. 

\section{Conclusion}\label{sec:conclusion}
In this paper, we exploited the fact that a practical form of Gabor functions is also a positive-definite kernel to find an image representation based on Gabor functions. 
This representation is learned by the stabilized infinite kernel learning regression algorithm  that had  been previously proposed by \citet{GhiasiShirazi2011}. 
The obtained representation has the weakness that the learned Gabor kernels are not localized and are present at all pixels.
We proposed a sparse representation algorithm based on LASSO and showed that in simple cases it can recover the underlying generating Gabor functions of images. 
As an application of our method, we proposed an algorithm for automatic learning of parameters of Gabor filters in the task of face recognition. 
Our experiments on CMU-PIE and Extended Yale B datasets confirm the usefulness of the proposed algorithm in automatic learning of Gabor filters.

\section{Acknowledgment}
The author wishes to express appreciation to Research Deputy of Ferdowsi University of Mashhad for supporting this project by grant No.: 2/38449. 
The author thanks Chuan-Xian Ren for providing him with the code of the MOST algorithm \citep{ren2014band} and the processed versions of CMU-PIE and Extended Yale B datasets. 
The author also thanks his colleagues, Ahad Harati and Ehsan Fazl-Ersi for their valuable comments.

\bibliography{GaborKernel}

\appendix

\section{Gabor functions as positive definite kernels}\label{sec:gabor-form}

Two dimensional Gabor functions have been studied in depth by \citet{lee1996image}. 
He starts from a general form of Gabor functions that consists of 8 parameters as follows:

\begin{eqnarray}\label{eq:gabor-general-form}
\begin{aligned}
\Psi&(x,y;x_0,y_0, \xi_0,\nu_0,\rho ,\theta,\sigma,\beta )=  \\
& \frac{1}{\sqrt{\pi\sigma\beta}}
e^{-\left(\frac{\left((x-x_0 )  cos{\theta}+(y-y_0 )  sin{\theta} \right)^2}{2\sigma^2 }
+\frac{\left(-(x-x_0 )  sin{\theta}+(y-y_0 )  cos{\theta} \right)^2}{2\beta ^2 }\right) } \\
& \times e^{i(\xi _0 (x-x_0 )+\nu_0 (y-y_0 )+\rho )}
\end{aligned}
\end{eqnarray}

\noindent
where the pair $(x_0,y_0)$ is the center of the filter in spatial domain, the parameters $\beta,\sigma,\theta$ determine an elliptical Gaussian, the parameters $\xi_0,\nu_0$ are the horizontal and vertical frequencies, and $\rho$ is the phase parameter. 
He then simplifies the form of Gabor functions by setting $\rho=0$.
Note that, even after setting $\rho=0$, the above form of Gabor functions is unwantedly too general and includes both Gaussian filters (when frequency parameters are zero) and sinusoidal waves (when $\sigma,\beta\to\infty$).
Considering the biological observations reported about the biological visual cells, \citet{lee1996image} reduces the number of parameters one by one until he arrives at a form with only two parameters. Similar two parameter forms for Gabor functions have been used by other researchers\citep{liu2002gabor,liu2004gabor,ren2014band, haghighat2013identification,saremi2013double}. All of these forms are special cases of  Eq.~(\ref{eq:gabor-general-form}) with $\rho=0$.
We now prove that Gabor functions are positive definite kernels.

\begin{proposition}
The complex-valued Gabor function $k([x,y],[x',y'])$ defined by the following equation is a positive definite kernel:

\begin{eqnarray}\label{eq:gabor-general-p.d-form}
\begin{aligned}
k(&[x,y],[x',y'])= \\ 
& \frac{1}{\sqrt{\pi\sigma\beta}}
e^{-\left(\frac{\left((x-x' )  cos{\theta}+(y-y' )  sin{\theta} \right)^2}{2\sigma^2 }
+\frac{\left(-(x-x' )  sin{\theta}+(y-y' )  cos{\theta} \right)^2}{2\beta ^2 }\right) } \\
& \times e^{i(\xi_0 (x-x' )+\nu_0 (y-y' ))}
.
\end{aligned}
\end{eqnarray}

\end{proposition}
\proof
{
Since the leading coefficient $\frac{1}{\sqrt{\pi\sigma\beta}}$ is positive and the class of positive definite kernels is closed under multiplication, it is enough to prove that the following functions are positive definite:

\begin{eqnarray}
\begin{aligned}
k_1(&[x,y],[x',y'])=\\
&\exp{\left(-\frac{\left((x-x' )  cos{\theta}+(y-y' )  sin{\theta} \right)^2}{2\sigma^2 }\right) }\\
k_2(&[x,y],[x',y'])=\\
&\exp{\left(-\frac{\left(-(x-x' )  sin{\theta}+(y-y' )  cos{\theta} \right)^2}{2\beta ^2 }\right) }\\
k_3&([x,y],[x',y'])= e^{i(\xi_0 (x-x' )+\nu_0 (y-y' ))}
.
\end{aligned}
\end{eqnarray}

The functions $k_1$ and $k_2$ are positive definite since they are Gaussian functions with general covariance matrices. 
Considering the fact that any function of the form $k(z,z') = f(z) \bar{f(z')}$ is positive definite\citep{Taylor_2004},
positive definiteness of $k_3$ follows from the following equation: 

\begin{eqnarray}
\begin{aligned}
k_3&([x,y],[x',y'])= e^{i(\xi_0 x+\nu_0 y)}e^{-i(\xi_0 x'+\nu_0 y')} 
\end{aligned}
\end{eqnarray}
.
\noindent
}

\begin{corollary}[]
Any real-valued Gabor function expressible in the following form is a positive definite kernel.

\begin{eqnarray}\label{eq:real-gabor-general-p.d-form}
\begin{aligned}
k(&[x,y],[x',y'])=  \\
& \frac{1}{\sqrt{\pi\sigma\beta}}
e^{-\left(\frac{\left((x-x')  cos{\theta}+(y-y')  sin{\theta} \right)^2}{2\sigma^2 }
+\frac{\left(-(x-x')  sin{\theta}+(y-y')  cos{\theta} \right)^2}{2\beta ^2 }\right) } \\
& \times cos(\xi_0 (x-x')+\nu_0 (y-y'))
\end{aligned}
\end{eqnarray}
.

\end{corollary}
\proof
{
	This follows from the fact that the real part of a complex-valued positive definite kernel function is a real-valued positive definite kernel \citep[see][page 31]{Scholkopf_2002}.
}

\end{document}